\documentclass[10pt]{article} % For LaTeX2e
\usepackage[preprint]{tmlr}

\usepackage{amsmath,amsfonts,bm}

\def\eqref#1{equation~\ref{#1}}
\def\1{\bm{1}}

\DeclareMathAlphabet{\mathsfit}{\encodingdefault}{\sfdefault}{m}{sl}
\SetMathAlphabet{\mathsfit}{bold}{\encodingdefault}{\sfdefault}{bx}{n}

\usepackage{booktabs}
\usepackage{hyperref}
\usepackage{url}
\usepackage{multirow}
\usepackage{graphicx}

\title{Impact of Language Guidance: A Reproducibility Study}

\author{%
\name Cherish Puniani\thanks{Equal contribution.} \email cherish\_p@me.iitr.ac.in \\
\addr Department of Mechanical Engineering\\
Indian Institute of Technology, Roorkee
\AND
\name Advika Sinha\footnotemark[1] \email advika\_s@me.iitr.ac.in \\
\addr Department of Mechanical Engineering\\
Indian Institute of Technology, Roorkee
\AND
\name Shree Singhi\footnotemark[1] \email shree\_s@mfs.iitr.ac.in\\
\addr Department of Data Science \& Artificial Intelligence\\
Indian Institute of Technology, Roorkee
\AND
\name Aayan Yadav\footnotemark[1] \email aayan\_y@mfs.iitr.ac.in\\
\addr Department of Data Science \& Artificial Intelligence\\
Indian Institute of Technology, Roorkee
}

\begin{document}
\maketitle

\newcommand{\fix}{\marginpar{FIX}}
\newcommand{\new}{\marginpar{NEW}}

\def\month{MM}  % Insert correct month for camera-ready version
\def\year{YYYY} % Insert correct year for camera-ready version
\def\openreview{\url{https://openreview.net/forum?id=XXXX}} % Insert correct link to OpenReview for camera-ready version

\maketitle

\begin{abstract}
Modern deep-learning architectures need large amounts of data to produce state-of-the-art results. Annotating such huge datasets is time-consuming, expensive, and prone to human error. Recent advances in self-supervised learning allow us to train huge models without explicit annotation. Contrastive learning is a popular paradigm in self-supervised learning. Recent works like SimCLR and CLIP rely on image augmentations or directly minimizing cross-modal loss between image and text. \cite{banani2023learning} propose to use language guidance to sample view pairs. They claim that language enables better conceptual similarity, eliminating the effects of visual variability. We reproduce their experiments to verify their claims and find that their dataset, RedCaps, contains low-quality captions. We use an off-the-shelf image captioning model, BLIP-2, to replace the captions and improve performance, and we also devise a new metric to evaluate the semantic capabilities of self-supervised models based on interpretability methods.
\end{abstract}

\section{Introduction}

Deep learning thrives on large datasets and compute-intensive training. While unlabeled data is abundant, supervised learning algorithms require annotated data. Annotation of huge datasets is prohibitively expensive, labour-intensive, and error-prone. Self-supervised learning (SSL) enables the model to learn rich and transferable representations from unlabeled data. This has unlocked new possibilities in both computer vision \citep{chen2020simple,caron2021emerging} and natural language processing \citep{devlin2018bert}.

Contrastive learning is a self-supervised learning technique in which a model is trained to bring similar images near by in embedding space while pushing dissimilar images far away. SimCLR \citep{chen2020simple} uses image augmentations such as random crop, Gaussian blur, and random flipping to generate a positive pair while treating other images as negative samples. Other methods \citep{caron2018deepcluster, wu2018unsupervised} use clustering algorithms or nearest neighbour operations to find positive samples. These methods only use visual similarity to find similar images. Two objects might be visually similar, while objects of the same class might be visually dissimilar. In contrast to this, conceptually similar images are more often described similarly. This suggests that leveraging language modality can improve contrastive learning.

\citet{radford2021learning} proposes learning a joint embedding space for images and their captions. This yields highly generalizable and accurate representations. However, \citet{banani2023learning} suggests that combining embedding spaces of different modalities might lead to sub-optimal results. They propose a new sampling procedure for contrastive learning where image pairs are sampled using caption similarity based on embeddings generated using a pre-trained language encoder.

\citet{banani2023learning} retrain existing self-supervised visual learning architectures \citep{chen2020simple,caron2018deepcluster,wu2018unsupervised} with the proposed sampling strategy. Their experiments show that the newly proposed method outperforms all baselines on varying downstream tasks across multiple datasets. This substantiates the claim that language is a good proxy for conceptual similarity.

We aim to rigorously evaluate these claims by closely replicating the experimental setup and results reported in the original paper. We identify the poor caption quality of the dataset\citep{desai2021redcaps} used by the original authors and generate better captions from an off-the-shelf caption generator \citep{li2023blip2} and analyze the performance improvement. We also demonstrate that the model learns semantic information using interpretability methods \citep{selvaraju2017grad}.

\begin{figure}
    \centering
    \includegraphics[width=1\linewidth]{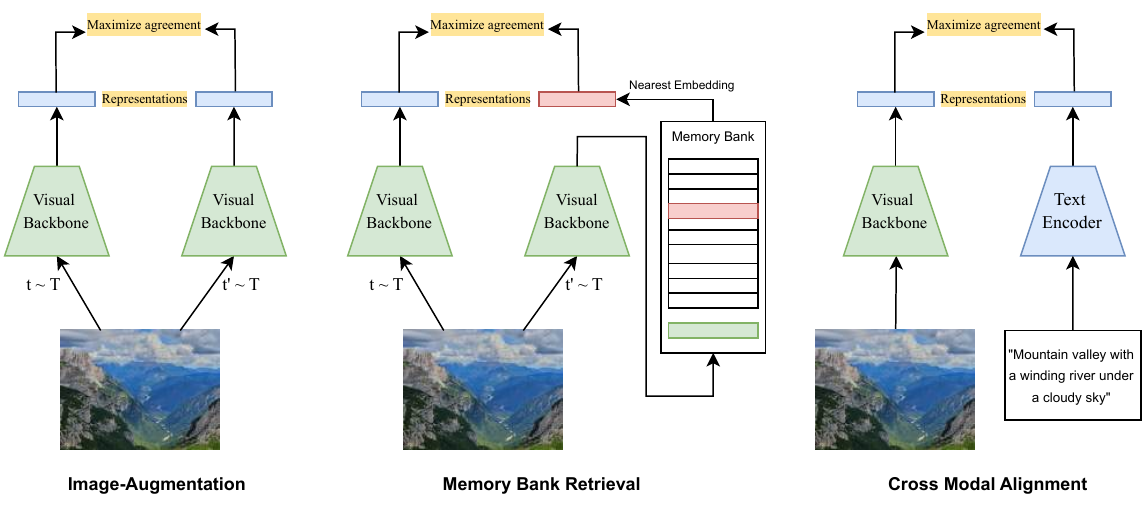}
    \caption{\textbf{Refining Representation Learning.} While early contrastive learning methods relied on simple image transformations, newer structured retrieval techniques have emerged to refine the learned embeddings beyond instance-level. These employ clustering, memory banks, or language-driven sampling to introduce structure into training signals and subsequently improve visual representation learning. }
    \label{fig:enter-label}
\end{figure}
\section{Scope for Reproducibility}

The main contribution of the original paper is a new language-based sampling strategy, and their claim is that this strategy improves the underlying self-supervision framework \citep{banani2023learning}.

In an effort to reproduce the paper and gain a deeper understanding, we discovered several key limitations:
\begin{itemize}   
    \item \textbf{Inefficient Captions:} The method’s efficacy is heavily dependent on image captions. Manual inspection of the dataset, RedCaps, reveals that the captions scraped from Reddit are often noisy, vague and inaccurate potentially hampering the model training process.
    
    % \item \textbf{RedCaps Dependency:} The method achieves optimal performance when pairs are sampled from specific subreddit subsets, introducing weak supervision \citep{zheng2021weakly}. The dataset-specific constraints reduce the generalizability of the method.
    
    \item \textbf{RedCaps Dependency:} The method relies on the availability of captions for the images. It achieves optimal performance when pairs are sampled from specific subreddit subsets. This introduces weak supervision \citep{zheng2021weakly}. The dataset-specific constraints reduce the generalizability of the method.

    \item \textbf{High Computational Requirements:} The reported results utilized ResNet-50\citep{he2016deep} with a batch size of 512 trained from scratch, which requires substantial computational resources that may not be readily available in many settings.
    
\end{itemize}

\textbf{Our Contributions} To address these limitations and extend the work, we make the following contributions:
\begin{itemize}
    \item \textbf{Reproducibility:} We provide an exhaustive replication of most experiments in the original paper, adapted for reduced computational environments.
    
    \item \textbf{Visual Backbone Optimization:} We investigate the generalization capabilities of language-guided SSL to smaller, more efficient architectures such as ResNet34 \citep{wightman2021resnet} and MobileNetV3 \citep{howard2017mobilenets}, making the approach accessible within resource constraints.
    
    \item \textbf{Caption Quality Enhancement:} We develop a curated set of refined captions for the existing dataset \citep{desai2021redcaps}, improving the impact of language guidance and reducing dataset dependency and enabling generalization to diverse datasets.
    
    % \item \textbf{Weak Supervision Analysis:} We conduct a comprehensive study on the impact of subreddit sub-class weak supervision on model performance \citep{zheng2021weakly}.
    
    \item \textbf{New Metric:} We generate saliency maps\citep{selvaraju2017grad}, which are used to create a new metric for evaluating SSL-trained ConvNets.
\end{itemize}

\section{Background}
In this section, we provide an overview of the foundational works and existing research that have contributed to advancements in this field. We discuss relevant literature, methodologies, and key developments that form the basis of our study, highlighting their significance in the context of our work.

\textbf{Visual Representation Learning} involves learning to encode visual information in an embedding space that preserves its semantics well. Unlike typical machine learning tasks like classification or segmentation, we cannot manually annotate ground truth labels for this task. So, we cannot directly optimize loss in embedding space. Two main approaches have been explored for this task: generative and discriminative. Generative approaches \citep{doersch2015unsupervised,gidaris2018unsupervised,oord2018representation,vincent2008extracting,zhang2016colorful} involve learning a model that can capture image distribution well. Such models are hypothesized to learn semantically relevant features. Discriminative approaches involve learning a model that can differentiate between images. Understanding semantically relevant features is essential for excelling in tasks like metric learning \citep{chopra2005learning}, dimensionality reduction \citep{hadsell2006dimensionality} and classification \citep{sharif2014cnn}. Self-supervised learning has recently gained popularity as a visual representation learning method. They relieve the need for human annotation and allow learning from large, unlabelled data sources. Various contrastive \citep{wu2018unsupervised,chen2020simple,chen2020big,chen2020mocov2,he2020moco}, and non-contrastive \citep{chen2021simsiam,grill2020bootstrap} approaches have been proposed recently. \citep{banani2023learning} propose a sampling strategy for contrastive visual representation learning.

\textbf{Image-Image Contrastive Learning.} Contrastive learning involves learning an embedding space in which similar images are close and dissimilar images are far away. Sampling positive and negative pairs effectively is an essential task for the effectiveness of contrastive learning. \citet{chen2020simple} propose a framework called SimCLR for contrastive learning. It involved using data augmentation to generate positive samples for each instance while treating all other images in the batch as negative samples. SimSiam, proposed by \citep{chen2021simsiam}, uses a similar approach but does not use negative pairs during training. It also implements a stop gradient operation in one branch of the Siamese network. It claims that the stop gradient operation is essential to prevent model collapse. \citet{caron2020unsupervised} propose SwAV, a new method which reduces computational complexity as it does not need to calculate explicit pairwise comparisons. It relies on an online clustering approach. They compute the feature of an image and then compute its code by matching it with a set of $k$ prototype vectors. Then, they predict the code from an augmented view of the image. These approaches mitigate the demand for a memory bank and reduce computational complexity, ignoring similarities between different instances. NNCLR, proposed by \citet{dwibedi2021nnclr}, utilizes similarity between different instances along with transformations to account for more semantic variation. It uses a memory bank similar to \citet{he2020moco} and samples the nearest neighbour of the image in latent space. It then minimizes the loss between the nearest neighbour and random augmentation of the original image.

\textbf{Using language for contrastive learning} 
Past work has aimed to learn joint vision-language representations for tasks like Visual Question Answering \citep{antol2015vqa,goyal2017making,hudson2019gqa,zhu2016visual7w}, Visual Reasoning \citep{kazemzadeh2014referitgame,suhr2019corpus,zellers2019recognition} and Retrieval \citep{park2022normalized,young2014image}. \citet{radford2021learning} introduced CLIP, which learned a joint vision-language embedding space by directly minimizing cross-modal loss. This approach was widely adopted due to its generalization and few shot capabilities. Other work improved upon this by adding additional losses and other improvements \citep{jia2021scaling,xu2022groupvit,yao2022filip,cui2022democratizing,lee2022uniclip,li2022declip,mu2021slip}. 

\citet{banani2023learning} aims to achieve good results in image-image contrastive learning with the help of language to sample positive pairs.

\begin{figure}
    \centering
    \includegraphics[width=0.8\linewidth]{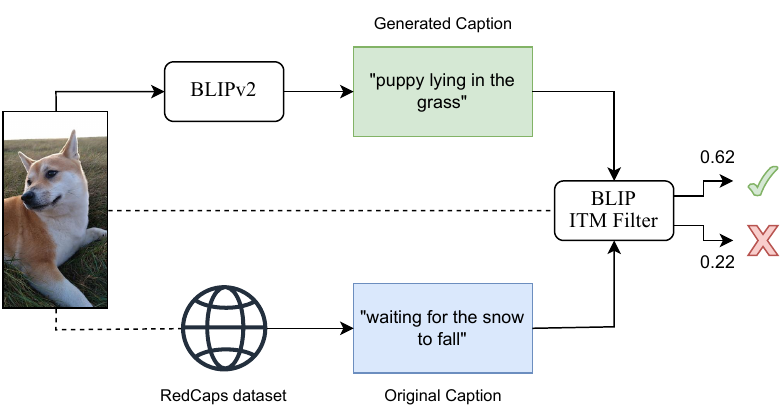}
    \caption{\textbf{Improving Captions via Contrastive Filtering.} Our caption improvement method leverages BLIPv2 to generate candidate captions that better describe an image. Since dataset-provided captions can be less relevant or inaccurate, we introduce an Image-Text Matching (ITM) Filter to evaluate, assign a relevance score and select the most appropriate caption between of the two. This ensures better semantic and visual alignment of a caption with its corresponding image.}
    \label{fig:capfilt}
\end{figure}

\section{Methodology}
Traditional SSL methods rely on image augmentations like cropping, colour jitter and noise addition to generate positive pairs. The underlying hypothesis is that visual similarity mirrors semantic similarity. While these methods learn features invariant to specific augmentations, they fail to capture semantic similarity well.

Our reproducibility study examines a language-guided sampling strategy that uses textual captions to identify semantically similar images. The hypothesis is that similar captions capture shared conceptual content beyond what visual augmentations can provide.

Most SSL frameworks use visual backbones like ConvNets and ViTs \citep{caron2021emerging} to learn visual representations.
\cite{banani2023learning} use ResNet50 as the visual backbone for their experiments. To test the generalizability of the framework we performed all experiments for ResNet34. Although both have similar architectures and have roughly the same number of parameters, $25.5$M and $21.8$M, respectively, they differ significantly in terms of their feature embedding size, $2048$ and $512$, respectively.

\subsection{Pair Sampling}
\citet{banani2023learning} use RedCaps \citep{desai2021redcaps}, a dataset scraped from Reddit. It consists of images with user-written captions.
% The pair for each image is found by comparing the vector embedding of its caption to the rest of the dataset.
They make image pairs by comparing the similarity of their corresponding caption embeddings.
The embeddings are generated using SBERT \citep{reimers2019sbert} and the two captions with the greatest cosine similarity are paired.

Metrics like BLEU \citep{papineni2002bleu} and CIDER \citep{vedantam2015cider} are n-gram-based methods traditionally used to find caption similarity. However, n-gram-based approaches are sensitive to variations in phrasing and sentence structure. Though SPICE \citep{anderson2016spice} uses parse trees and handles structural variations better, it is still limited in dealing with different word choices for the same concept. The choice of right caption similarity metric is foundational to the success of their method. SBERT effectively identifies semantically similar caption pairs while being robust to surface-level text variations. The use of cosine similarity simplifies the implementation while maintaining reliable semantic matching capabilities. It is observed that the method is agnostic to the sentence encoder chosen.

The FAISS algorithm \citep{johnson2019faiss} is used to find the nearest neighbour for a caption in language embedding space. The image corresponding to the most similar caption is chosen as a positive sample for the original image. The similarity search took approximately 21 minutes to complete on our system. 

% Captions curated from Reddit are known to be very vague and noisy which we hypothesized hurts the performance of this method so we use BLIPv2  to generate short, descriptive captions for each image. This ensures consistent textual quality and alleviates the dependency on availability of captions in the dataset, it allows us to scale this approach to larger uncaptioned larger datasets which improves accuracy of all SSL methods in general.(fig. x)

\subsection{Improving the Dataset}
We identified that the quality of Reddit-sourced captions could be a significant limiting factor. The RedCaps dataset, while extensive, contains captions that are often vague, noisy, and inconsistent in their descriptive quality. To test this hypothesis and potentially improve the method, we introduced BLIPv2 as an alternative caption generation approach.
BLIPv2 generates concise, descriptive captions that maintain consistent quality across the dataset. We adopt a filtering strategy where we generate new captions using BLIPv2 and evaluate their quality using Image-Text Matching (ITM) scores. The ITM score measures the alignment between an image and its caption by predicting whether they are a good match. Higher ITM scores indicate captions that are more semantically relevant to the image. Captions with higher ITM scores are retained to ensure better language guidance for contrastive learning, as shown in \autoref{fig:capfilt}. The strategy can also be adopted in cases where captions are unavailable. In such cases, multiple captions can be sampled using BLIP and the caption with the highest ITM score can be retained.

Our modification serves two key purposes. Firstly, it allows us to evaluate whether higher-quality captions improve the performance of language-guided SSL. Secondly, it enables language guidance frameworks to be extended to any image dataset, regardless of whether it contains associated text descriptions.

\subsection{Visualizing learned representations}
\label{subsec:saliencygen}

We use self-supervised learning methods to learn visual representations. These representations are used for downstream tasks and evaluated on it. However, it is essential to inspect if the model is focussing on the right regions of the image. We train a linear probe on the ResNet-34 backbone using the train set of ImageNetS-50. We apply GradCAM \citep{selvaraju2017grad} on the second convolutional block of layer 4 of the backbone with true class of the image.

\begin{figure}
    \centering
    \includegraphics[width=1\linewidth]{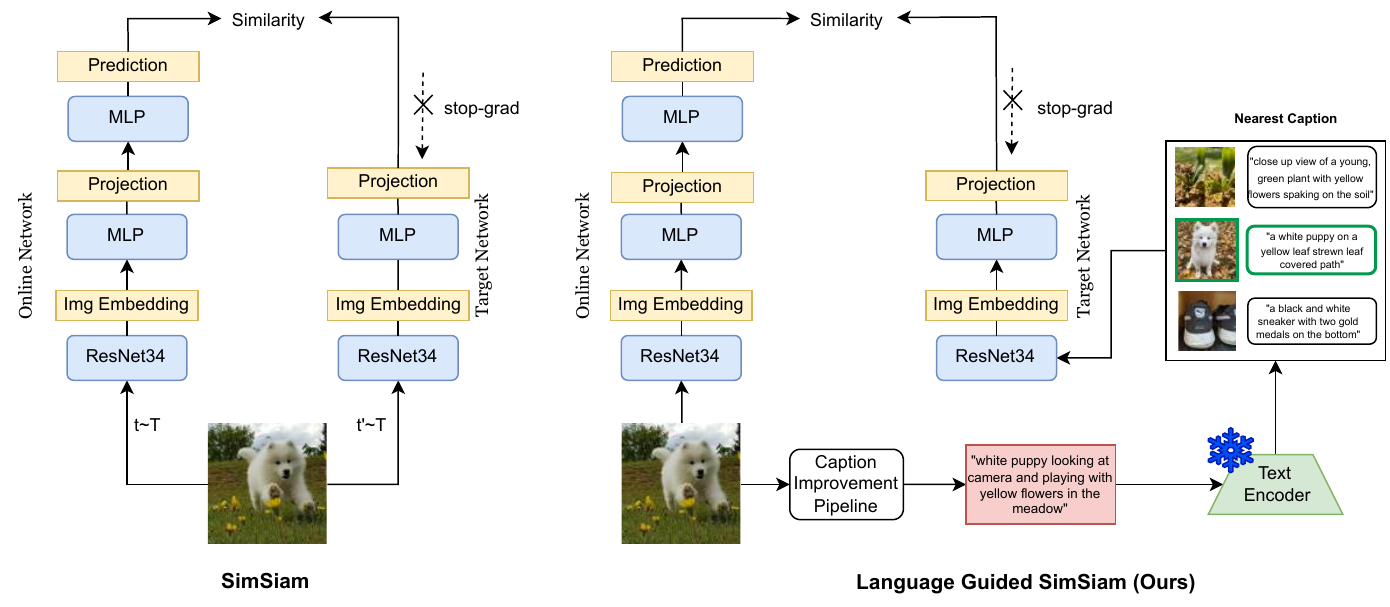}
    \caption{A schematic comparison of SimSiam and Language Guided SimSiam trained using our pipeline.}
    \label{fig:siamsiam}
\end{figure}

\section{Experimental Setup}
We primarily base our experiments on the code\footnote{https://github.com/mbanani/lgssl} provided by the authors. Our experimental evaluation focuses on thoroughly validating the impact of improved captions on self-supervised learning frameworks. We explore multiple frameworks while maintaining consistent training conditions across all experiments to ensure fair comparisons.

To comprehensively evaluate the effect of improved captions, we conduct experiments across a diverse set of self-supervised learning frameworks. Our study includes SimCLR, LGSimCLR, SimSiam, SwAV, and NNCLR. This selection enables us to verify whether the performance improvements from better captions generalize across different architectural approaches, as demonstrated in Table~\ref{tab:linearprobe}.

\subsection{Training Details}
We perform all experiments using a ResNet-34 backbone trained across on two NVIDIA V100s. To ensure meaningful comparisons across different experimental conditions, we maintain consistent training parameters throughout our studies. We employ the AdamW optimizer \citep{loshchilov2016sgdr} with the same hyperparameters used by \citet{banani2023learning}: learning rate of 0.001 and weight decay of 0.01. The learning schedule follows a cosine decay pattern with 5000 warm-up steps. Each model is trained for $25$ epochs with a batch size of $512$ images resulting in a training time of approximately 1.5 hours per epoch.

We choose RedCaps-2020 as our training dataset. It is a subset of the RedCaps dataset comprising 2.8 million image-text pairs uploaded on Reddit in 2020. We explore two distinct caption sources in our experiments. First, we establish baseline performance using the original RedCaps captions. Then, we use our caption enhancement pipeline to measure the direct impact of caption quality on model performance.

% To validate the generalizability of our findings, we extend our experiments to include a subset of the LAION-400M dataset. This additional evaluation, detailed in Table~\ref{tab:dataset_comparison}, helps demonstrate that our improvements are not specific to the RedCaps dataset characteristics.

\begin{figure}[h]
    \centering
    \includegraphics[width=1\linewidth]{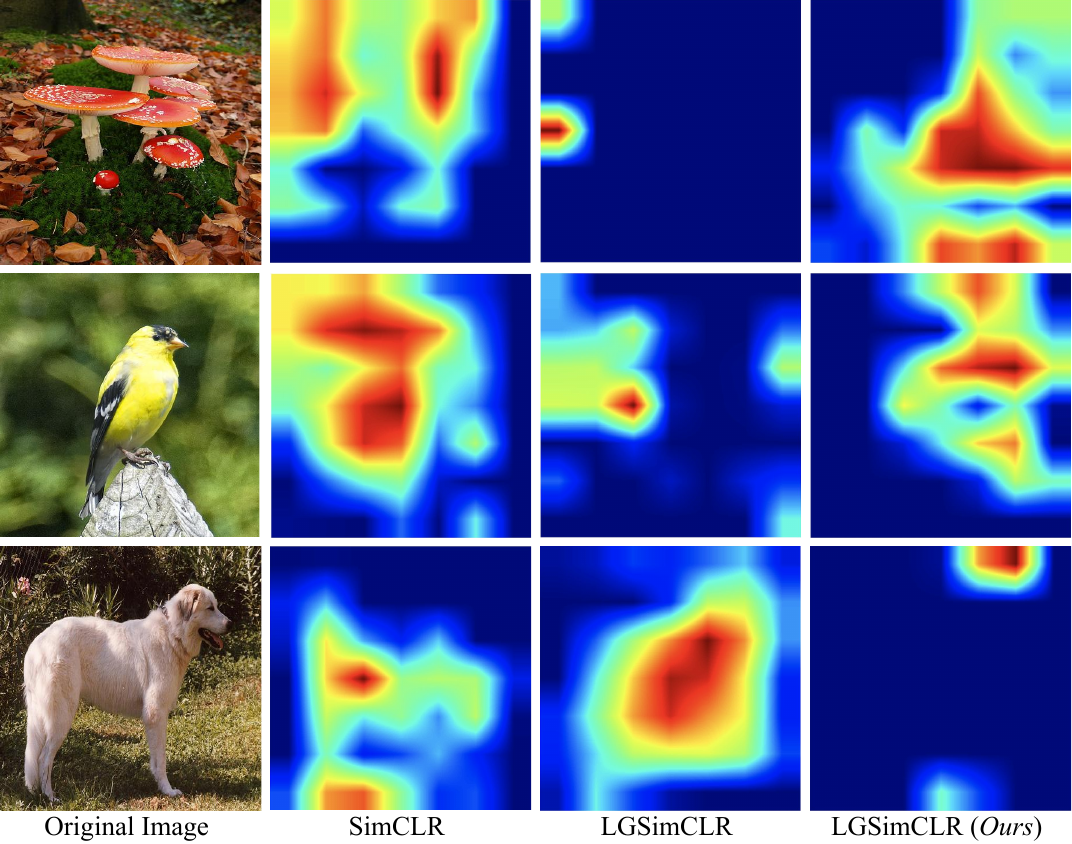}
    \caption{\textbf{Visualisations.} Different models perform better on different classes. \textbf{Top}: LGSimCLR (\textit{Ours}) performs the best. \textbf{Middle}: SimCLR performs the best. \textbf{Bottom}: LGSimCLR performs the best.}
    \label{fig:viz}
\end{figure}

\subsection{Evaluation Protocol} \label{subsec:evaluation}
We run different downstream tasks on the frozen features for each model across multiple datasets to evaluate their performance. Similar to the original authors, we evaluate the model on linear-probe classification \citep{kornblith2019better} and few-shot classification \citep{wang2019simpleshot}. We were able to reproduce results for all datasets mentioned in the original paper except Sun397, Cars, Caltech-101 and Oxford Flowers. The Sun397 dataset has several corrupted images; Cars dataset has been removed from the host site; and the authors' code implementation to download Caltech-101 and Oxford Flowers is not working. Additionally, we report results using a new approach to evaluate self-supervised models using saliency maps.

\begin{table}[h]
\centering
\begin{tabular}{lccc}
\hline
\textbf{Metric} & \textbf{SimCLR} & \textbf{LGSimCLR} & \textbf{LGSimCLR (\textit{Ours}) }\\
\hline
AUC-ROC & 0.5411 & 0.5195 & \textbf{0.5501}\\
AUC-PR & \textbf{0.3419} & 0.3244 & 0.3416\\
\hline
\end{tabular}
\caption{Saliency Map Evaluation. We evaluate across three models i.e. SimCLR, LGSimCLR on original captions and LGSimCLR on new captions (Ours) across two metrics. All the models were retrained with ResNet-34 backbone.}
\label{tab:sal_metrics}
\end{table}

\textbf{Saliency Map Evaluation}
We generate saliency maps using the method described in \autoref{subsec:saliencygen}. We evaluate the saliency maps using the Area Under the Precision-Recall Curve (AUC-PR) and the Area Under the ROC Curve (AUC-ROC) \citep{cong2018review}. These are calculated by treating the ground truth segmentation map as targets for pixel-level classification. We use the validation set of ImageNet-S50 for evaluation. Results are reported in \autoref{tab:sal_metrics}. We can see that the metrics are not significantly different. So, language guidance does not seem to have a significant impact on the quality of saliency maps. We visualize saliency maps for all validation images and notice that each model performs better than others on some categories while performing worse on others. We display this for 3 out of 50 categories in \autoref{fig:viz}.

\section{Results and Discussion}
We report the classification results in \autoref{tab:linearprobe}, \autoref{tab:fewshot} and \autoref{tab:imagenet}. Our experiments suggest that the impact of language guidance is not as profound as indicated in the original paper. The performance disparity between ResNet34 and ResNet50 may be largely attributed to differences in the size of their feature embedding spaces since their architecture and overall parameter counts are comparable, $21.79$M and $25.5$M respectively. ResNet50 produces a $2048$-dimensional feature representation in contrast to the $512$ dimensional feature representation of ResNet34. The large embedding space of ResNet50 likely allows the network to capture a rich set of features. Conversely, the reduced capacity of a $512$-dimensional embedding space may limit the model's ability to fully exploit the semantic cues provided by the language guidance, resulting in a less pronounced improvement in performance. Neglecting this factor might have led to an overestimation of the generalizability of this method. We also ran experiments using MobileNetV3 as the backbone. The classification results were near-random across all models. We hypothesise that this was either due to the significantly small model size, $5.5$M parameters, or smaller embedding space of $1024$ dimensions.

%CHERISH
% Additionally, our caption improvement approach strengthens language guidance and improves performance across most datasets frameworks by providing clearer, more consistent captions. This reduces ambiguity and allows models to align visual features with concepts better, resulting in improved performance.
Our caption improvement pipeline performs better than \cite{banani2023learning} in most cases. It boosts LGSimSiam's performance on the 10 fine-grained datasets and few-shot evaluation on ImageNet but fails to show improvement on linear probe evaluation on ImageNet. Similarly, our pipeline improves SimCLR's performance on ImageNet and few-shot evaluation on the fine-grained datasets but slightly worsens the performance on linear probe evaluation for the fine-grained dataset. This inconsistency could be attributed to the much-reduced output embedding size of the ResNet34 causing stochasticity in results.

%this upar wala part seems thoda harsh on us, kaise change karu?
%idk -Aayan? Wbu @SHREE?

\begin{figure}
        \centering
        \includegraphics[width=1\linewidth]{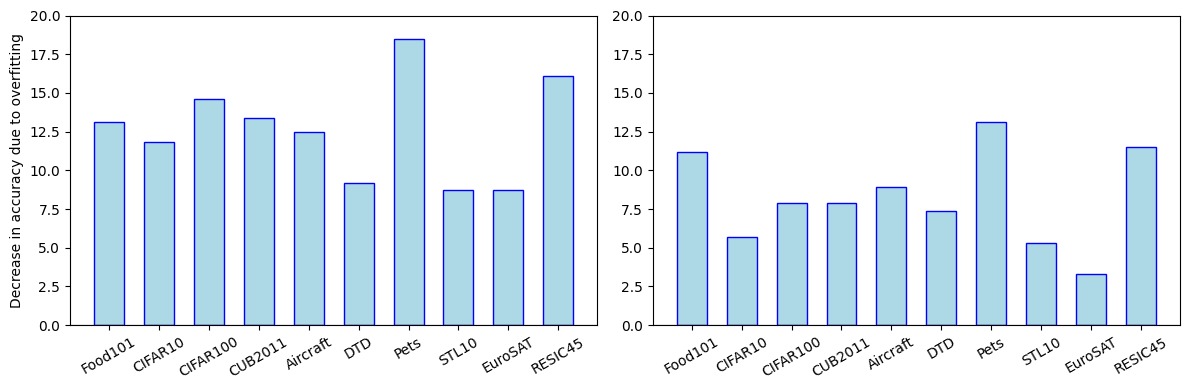}
        \caption{\textbf{Overfitting of LGSimSiam.} We visualize the advantage of using early stopping by comparing the performance of LGSimSiam at the $25^{th}$ and $6^{th}$ epoch. \textbf{Left:} \citet{banani2023learning} \textbf{Right:} Ours.}
    \label{fig:overfit-simsiam}
\end{figure}

\begin{figure}
    \centering
    \includegraphics[width=1\linewidth]{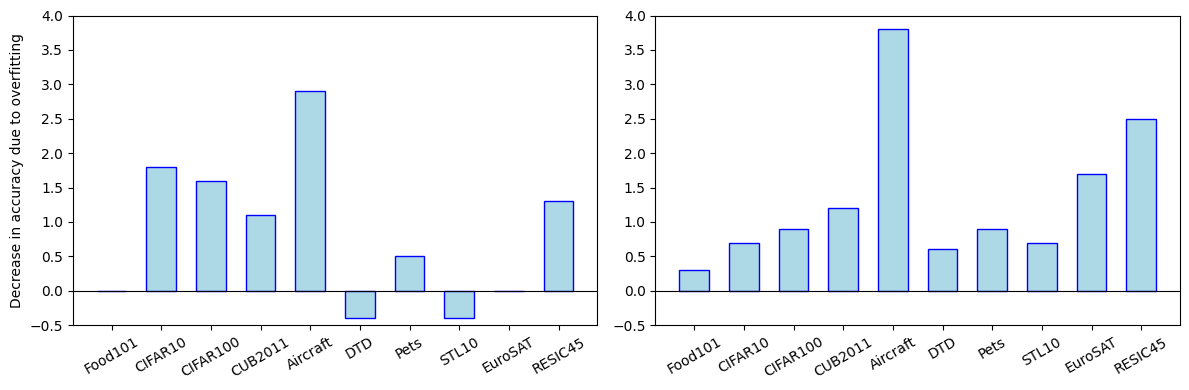}
    \caption{\textbf{Overfitting of LGSimCLR.} We visualize the advantage of using early stopping by comparing the performance of LGSimCLR at the $25^{th}$ and $9^{th}$ epoch. \textbf{Left:} \citet{banani2023learning} \textbf{Right:} Ours.}
    \label{fig:overfit-simclr}
\end{figure}

Visual baseline models showed a gradual improvement till the end of the training. In contrast, the language-guided counterparts began to overfit soon, hence, performance significantly worsened as shown in \autoref{fig:overfit-simclr} and \autoref{fig:overfit-simsiam}. Based on these observations, we report results at the epoch with the highest validation performance for language guided SimCLR and SimSiam at the $9^{th}$ and $6^{th}$ epoch, respectively.

% \begin{table}[h]
% \setlength{\tabcolsep}{2pt}
% \centering
% \small
% \begin{tabular}{lccccccccccc}
% \hline
% \textbf{Models} & \textbf{Food101} & \textbf{CIFAR10} & \textbf{CIFAR100} & \textbf{CUB} & \textbf{Aircraft} & \textbf{DTD} & \textbf{Pets} & \textbf{STL10} & \textbf{EuroSAT} & \textbf{RESISC45} & \textbf{Avg} \\
% \hline
% SimCLR &61.2&77.5&53.0&27.3&36.0&61.9&66.4&85.1&94.2&78.7&\textbf{64.13}\\
% VisSimSiam &56.0&72.0&46.1&21.1&27.9&58.6&55.9&85.2&92.3&72.4&58.75\\
% VisNNCLR & 52.7& 69.9&45.4 &17.3 &25.7 & 59.2& 56.8&83.6 & 91.8& 71.8&57.42\\
% SWAV &50.7 &69.0 &43.0 &15.3 &23.2 &57.6 &54.1 &83.2 &90.4 &68.6 &55.51\\
% \hline
% \textit{Banani et al.} \\
% \hline
% LGSimCLR &66.7 &72.9 &51.1 &38.5 &32.4 &59.3 &65.5 &84.2 &89.8 &77.5 &63.79\\
% LGSimSiam &54.9 &71.9 &49.8 &23.5 &30.7 &52.6 &46.9 &79.0 &90.1 &75.1 &57.45\\
% \hline
% \textit{Ours} \\
% \hline
% LGSimCLR &64.4 &75.0 &51.2 &33.4 &30.7 &59.5 &67.8 &87.3 &91.1 &77.2 &63.76\\
% LGSimSiam &55.8 &73.8 &50.8 &23.6 &28.6 &56.1 &53.4 &85.7 &90.2 &74.8 &59.28\\  
% \hline
% \end{tabular}
% \caption{\textbf{Linear Probe.} We report performance of a linear probe using frozen features on 10 downstream tasks. LGSimCLR-\textit{Ours} outperforms
%  previous approaches on most datsets. \textit{Ours} refers to the models trained on new captions. We retrain all models with ResNet-34 backbone.}
% \label{tab:linearprobe}
% \end{table}

\begin{table}[h]
\setlength{\tabcolsep}{2pt}
\centering
\small
\begin{tabular}{lccccccccccc}
\hline
\textbf{Models} & \textbf{Food101} & \textbf{CIFAR10} & \textbf{CIFAR100} & \textbf{CUB} & \textbf{Aircraft} & \textbf{DTD} & \textbf{Pets} & \textbf{STL10} & \textbf{EuroSAT} & \textbf{RESISC45} & \textbf{Avg} \\
\hline
NNCLR & 52.7& 69.9&45.4 &17.3 &25.7 & 59.2& 56.8&83.6 & 91.8& 71.8&\textbf{57.42}\\
SWAV &50.7 &69.0 &43.0 &15.3 &23.2 &57.6 &54.1 &83.2 &90.4 &68.6 &55.51\\
\hline
\textit{SimCLR} \\
\hline
Visual &61.2&77.5&53.0&27.3&36.0&61.9&66.4&85.1&94.2&78.7&\textbf{64.13}\\
Banani et al. &66.7 &72.9 &51.1 &38.5 &32.4 &59.3 &65.5 &84.2 &89.8 &77.5 &63.79\\
Ours &64.4 &75.0 &51.2 &33.4 &30.7 &59.5 &67.8 &87.3 &91.1 &77.2 &63.76\\
\hline
\textit{SimSiam} \\
\hline
Visual &56.0&72.0&46.1&21.1&27.9&58.6&55.9&85.2&92.3&72.4&58.75\\
Banani et al. &54.9 &71.9 &49.8 &23.5 &30.7 &52.6 &46.9 &79.0 &90.1 &75.1 &57.45\\
Ours &55.8 &73.8 &50.8 &23.6 &28.6 &56.1 &53.4 &85.7 &90.2 &74.8 &\textbf{59.28}\\  
\hline
\end{tabular}
\caption{\textbf{Linear Probe Classification.} We report the performance of a linear probe using frozen features on 10 downstream tasks. The first split refers to models that language guidance is not applicable to. The second and third splits refer to different versions of SimCLR and SimSiam respectively.}
\label{tab:linearprobe}
\end{table}

\begin{table}[h]
\setlength{\tabcolsep}{2pt}
\centering
\small
\begin{tabular}{lccccccccccc}
\hline
\textbf{Models} & \textbf{Food101} & \textbf{CIFAR10} & \textbf{CIFAR100} & \textbf{CUB} & \textbf{Aircraft} & \textbf{DTD} & \textbf{Pets} & \textbf{STL10} & \textbf{EuroSAT} & \textbf{RESISC45} & \textbf{Avg} \\
\hline
NNCLR &64.0 & 50.9& 56.4& 45.1& 33.8&70.7 & 71.0&74.8 & 75.2&69.3 &\textbf{61.12}\\
SWAV &64.5 &51.6 &55.8 &44.0 &33.9 &71.2 &69.6 &74.3 &72.3 &68.5 &60.57\\
\hline
\textit{SimCLR} \\
\hline
Visual &67.8&52.9&59.5&54.7&41.5&74.6&73.6&73.9&82.0&77.5&65.80\\
Banani et al. &82.5 &51.6 &63.3 &73.4 &42.7 &73.5 &76.2 &75.2 &76.3 &82.5 &69.72 \\
Ours &80.2 &56.8 &66.1 &68.8 &39.5 &72.6 &78.1 &80.8 &78.5 &81.3 & \textbf{70.27}\\
\hline
\textit{SimSiam} \\
\hline
Visual &61.2&51.4&56.6&43.6&33.5&72.1&62.9&73.1&75.2&68.6&59.82\\
Banani et al. &70.5 &52.9 &62.4 &51.4 &38.6 &69.6 &55.7 &70.1 &76.6 &79.0 &62.68\\
Ours &72.4 &54.5 &64.8 &53.2 &36.8 &70.3 &61.2 &78.3 &77.9 &78.7 &\textbf{64.81}\\
\hline
\end{tabular}
\caption{\textbf{Few Shot Classification.} We report the performance of 5-way, 5-shot classification using frozen features on 10 downstream tasks. LGSimCLR-\textit{Ours} outperforms
 previous approaches on most datasets.}
\label{tab:fewshot}
\end{table}

% % GPT generated table bana raha hu
% \begin{table}[h]
% \setlength{\tabcolsep}{4pt}
% \centering
% \small
% \begin{tabular}{lcc}
% \hline
% \textbf{Model} & \textbf{Linear} & \textbf{Few-shot} \\
% \hline
% SimCLR & 68.5 & 82.2 \\
% SimSiam & 65.2 & 80.3 \\
% SwAV & 72.0 & 81.2 \\
% NNCLR & 66.1 & 80.9 \\
% \hline
% LGSimSiam & 65.8 & 83.7 \\
% LGSimSiam(ours) & 65.4 & 84.5 \\
% LGSimCLR & 67.0 & 87.8 \\
% LGSimCLR(Ours) & 72.7 & 88.1 \\
% \hline
% \end{tabular}
% \caption{Performance comparison of different models using linear and few-shot evaluation on imagenet.}
% \label{tab:performance}
% \end{table}
\begin{table}[h]
\setlength{\tabcolsep}{4pt}
\centering
\small
\begin{tabular}{lcccccccc}
\hline
\multirow{2}{*}{} & \multicolumn{4}{c}{\textit{Without language guidance }}&\multicolumn{2}{c}{\textit{Original Captions}}&\multicolumn{2}{c}{\textit{Our Captions}} \\ 
\cmidrule(lr){2-5} \cmidrule(lr){6-7} \cmidrule(lr){8-9}
 & \textbf{SimCLR} & \textbf{SimSiam} & \textbf{SwAV} & \textbf{NNCLR} & \textbf{LGSimSiam} &  \textbf{LGSimCLR}&\textbf{LGSimSiam}  & \textbf{LGSimCLR} \\
\hline

Linear & 68.5 & 65.2 & 72.0 & 66.1 & 65.8 & 67.0&65.4  & \textbf{72.7} \\
Few-shot & 82.2 & 80.3 & 81.2 & 80.9 & 83.7 & 87.8 &84.5 &  \textbf{88.1} \\
\hline
\end{tabular}
\caption{Performance comparison of different models using linear and few-shot evaluation on ImageNet.}
\label{tab:imagenet}
\end{table}

\section{Communication with Authors}
\citet{banani2023learning} conducted experiments on data subsampled from different subreddits. The code provided by them corresponding to these experiments was incomplete, lacking invocation of functions or definitions of functions themselves. We raised issues on their GitHub repository as it was unclear how to complete the code implementation. Attempts to resolve these roadblocks by raising issues in the author's GitHub repository were in vain, as the issues remained unaddressed.
 
\section{Limitations}
Our experiments, when compared to \cite{banani2023learning}, support the claim that incorporating a captioning model improves the impact of language guidance. However, the noisy and vague nature of the captions generated by the model limited the generalizability of our results. This underscores the need for a more accurate captioning model, which could potentially perform better in language-guided sampling tasks. The complete reproducibility of the original experiments could not be confirmed for three primary reasons. Firstly, the code implementation for subreddit subsampling was incomplete. Secondly, as mentioned in \autoref{subsec:evaluation}, certain datasets could not be included in our evaluative experiments. Additionally, when we substituted the ResNet backbone with a MobileNet variant, the performance notably worsened, highlighting the backbone dependency present in the original framework. 

\section{Conclusion}
Our study confirms that language guidance can improve self-supervised learning, but the effectiveness heavily depends on dataset quality, backbone selection, and training strategies. We show how early stopping is crucial when training with language guidance, as the models tend to overfit within a few epochs. Without this precaution, performance gains may be misleading or lost entirely. Additionally, dataset reliability plays a key role in the success of language guidance. Our evaluations revealed that low-quality captions can degrade performance, while improved captions generated using BLIP-2 led to more consistent results. This emphasizes the need for careful dataset curation when using language-based methods in contrastive learning.

Furthermore, our findings highlight the importance of embedding size in self-supervised learning. The performance gap between ResNet50 and ResNet34 suggests that larger embedding space is necessary for ConvNets to capture better semantic relationships. This is critical when choosing architectures for language-guided training.
Overall, our reproducibility study underscores both the potential and the limitations of language guidance in self-supervised learning. By identifying key factors like overfitting, embedding size, and dataset quality, we provide actionable insights for future research and real-world applications.

\bibliography{main}

\begin{thebibliography}{54}
\providecommand{\natexlab}[1]{#1}
\providecommand{\url}[1]{\texttt{#1}}
\expandafter\ifx\csname urlstyle\endcsname\relax
  \providecommand{\doi}[1]{doi: #1}\else
  \providecommand{\doi}{doi: \begingroup \urlstyle{rm}\Url}\fi

\bibitem[Anderson et~al.(2016)Anderson, Fernando, Johnson, and Gould]{anderson2016spice}
Peter Anderson, Basura Fernando, Mark Johnson, and Stephen Gould.
\newblock {SPICE: Semantic propositional image caption evaluation}.
\newblock In \emph{Proceedings of European Conference on Computer Vision (ECCV)}, 2016.

\bibitem[Antol et~al.(2015)Antol, Agrawal, Lu, Mitchell, Batra, Lawrence~Zitnick, and Parikh]{antol2015vqa}
Stanislaw Antol, Aishwarya Agrawal, Jiasen Lu, Margaret Mitchell, Dhruv Batra, C~Lawrence~Zitnick, and Devi Parikh.
\newblock {VQA}: Visual question answering.
\newblock In \emph{Proceedings of IEEE International Conference on Computer Vision (ICCV)}, 2015.

\bibitem[Banani et~al.(2023)Banani, Desai, Johnson, and Ghanem]{banani2023learning}
Mahdi Banani, Karan Desai, Justin Johnson, and Bernard Ghanem.
\newblock Learning visual representations via language-guided sampling.
\newblock In \emph{Proceedings of the IEEE/CVF Conference on Computer Vision and Pattern Recognition (CVPR)}, 2023.

\bibitem[Caron et~al.(2018)Caron, Bojanowski, Joulin, and Douze]{caron2018deepcluster}
Mathilde Caron, Piotr Bojanowski, Armand Joulin, and Matthijs Douze.
\newblock Deep clustering for unsupervised learning of visual features.
\newblock In \emph{Proceedings of the European conference on computer vision (ECCV)}, pp.\  132--149, 2018.

\bibitem[Caron et~al.(2020)Caron, Misra, Mairal, Goyal, Bojanowski, and Joulin]{caron2020unsupervised}
Mathilde Caron, Ishan Misra, Julien Mairal, Priya Goyal, Piotr Bojanowski, and Armand Joulin.
\newblock Unsupervised learning of visual features by contrasting cluster assignments.
\newblock In \emph{Advances in Neural Information Processing Systems (NeurIPS)}, volume~33, 2020.

\bibitem[Caron et~al.(2021)Caron, Touvron, Misra, J{\'e}gou, Mairal, Bojanowski, and Joulin]{caron2021emerging}
Mathilde Caron, Hugo Touvron, Ishan Misra, Herv{\'e} J{\'e}gou, Julien Mairal, Piotr Bojanowski, and Armand Joulin.
\newblock Emerging properties in self-supervised vision transformers.
\newblock In \emph{Proceedings of IEEE Conference on Computer Vision and Pattern Recognition (CVPR)}, 2021.

\bibitem[Chen et~al.(2020{\natexlab{a}})Chen, Kornblith, Norouzi, and Hinton]{chen2020simple}
Ting Chen, Simon Kornblith, Mohammad Norouzi, and Geoffrey Hinton.
\newblock {A Simple Framework for Contrastive Learning of Visual Representations}.
\newblock In \emph{Proceedings of the International Conference on Machine Learning (ICML)}, 2020{\natexlab{a}}.

\bibitem[Chen et~al.(2020{\natexlab{b}})Chen, Kornblith, Swersky, Norouzi, and Hinton]{chen2020big}
Ting Chen, Simon Kornblith, Kevin Swersky, Mohammad Norouzi, and Geoffrey~E Hinton.
\newblock Big self-supervised models are strong semi-supervised learners.
\newblock In \emph{Advances in Neural Information Processing Systems (NeurIPS)}, volume~33, 2020{\natexlab{b}}.

\bibitem[Chen \& He(2021)Chen and He]{chen2021simsiam}
Xinlei Chen and Kaiming He.
\newblock Exploring simple siamese representation learning.
\newblock In \emph{Proceedings of IEEE Conference on Computer Vision and Pattern Recognition (CVPR)}, 2021.

\bibitem[Chen et~al.(2020{\natexlab{c}})Chen, Fan, Girshick, and He]{chen2020mocov2}
Xinlei Chen, Haoqi Fan, Ross Girshick, and Kaiming He.
\newblock Improved baselines with momentum contrastive learning.
\newblock \emph{arXiv preprint arXiv:2003.04297}, 2020{\natexlab{c}}.

\bibitem[Chopra et~al.(2005)Chopra, Hadsell, and LeCun]{chopra2005learning}
Sumit Chopra, Raia Hadsell, and Yann LeCun.
\newblock Learning a similarity metric discriminatively, with application to face verification.
\newblock In \emph{Proceedings of IEEE Conference on Computer Vision and Pattern Recognition (CVPR)}, 2005.

\bibitem[Cong et~al.(2018)Cong, Lei, Fu, Cheng, Lin, and Huang]{cong2018review}
Runmin Cong, Jianjun Lei, Huazhu Fu, Ming-Ming Cheng, Weisi Lin, and Qingming Huang.
\newblock Review of visual saliency detection with comprehensive information.
\newblock \emph{IEEE Transactions on circuits and Systems for Video Technology}, 29\penalty0 (10):\penalty0 2941--2959, 2018.

\bibitem[Cui et~al.(2022)Cui, Zhao, Liang, Li, and Shao]{cui2022democratizing}
Yufeng Cui, Lichen Zhao, Feng Liang, Yangguang Li, and Jing Shao.
\newblock Democratizing contrastive language-image pre-training: A {CLIP} benchmark of data, model, and supervision.
\newblock In \emph{ICML Workshop on Pre-training: Perspectives, Pitfalls, and Paths Forward}, 2022.

\bibitem[Desai et~al.(2021)Desai, Kaul, Aysola, and Johnson]{desai2021redcaps}
Karan Desai, Gaurav Kaul, Zubin Aysola, and Justin Johnson.
\newblock {RedCaps: Web-curated image-text data created by the people, for the people}.
\newblock In \emph{NeurIPS Datasets and Benchmarks}, 2021.

\bibitem[Devlin et~al.(2018)Devlin, Chang, Lee, and Toutanova]{devlin2018bert}
Jacob Devlin, Ming-Wei Chang, Kenton Lee, and Kristina Toutanova.
\newblock Bert: Pre-training of deep bidirectional transformers for language understanding.
\newblock \emph{Proceedings of the Conference of the North American Chapter of the Association for Computational Linguistics -- Human Language Technologies (NAACL HLT)}, 2018.

\bibitem[Doersch et~al.(2015)Doersch, Gupta, and Efros]{doersch2015unsupervised}
Carl Doersch, Abhinav Gupta, and Alexei~A Efros.
\newblock Unsupervised visual representation learning by context prediction.
\newblock In \emph{Proceedings of IEEE International Conference on Computer Vision (ICCV)}, 2015.

\bibitem[Dwibedi et~al.(2021)Dwibedi, Aytar, Tompson, Sermanet, and Zisserman]{dwibedi2021nnclr}
Debidatta Dwibedi, Yusuf Aytar, Jonathan Tompson, Pierre Sermanet, and Andrew Zisserman.
\newblock With a little help from my friends: Nearest-neighbor contrastive learning of visual representations.
\newblock In \emph{Proceedings of the IEEE/CVF International Conference on Computer Vision (ICCV)}, pp.\  9588--9597, October 2021.

\bibitem[Gidaris et~al.(2018)Gidaris, Singh, and Komodakis]{gidaris2018unsupervised}
Spyros Gidaris, Praveer Singh, and Nikos Komodakis.
\newblock Unsupervised representation learning by predicting image rotations.
\newblock In \emph{Proceedings of the International Conference on Learning Representations (ICLR)}, 2018.

\bibitem[Goyal et~al.(2017)Goyal, Khot, Summers-Stay, Batra, and Parikh]{goyal2017making}
Yash Goyal, Tejas Khot, Douglas Summers-Stay, Dhruv Batra, and Devi Parikh.
\newblock Making the {V} in {VQA} matter: Elevating the role of image understanding in visual question answering.
\newblock In \emph{Proceedings of IEEE Conference on Computer Vision and Pattern Recognition (CVPR)}, 2017.

\bibitem[Grill et~al.(2020)Grill, Strub, Altch{\'e}, Tallec, Richemond, Buchatskaya, Doersch, Avila~Pires, Guo, Gheshlaghi~Azar, et~al.]{grill2020bootstrap}
Jean-Bastien Grill, Florian Strub, Florent Altch{\'e}, Corentin Tallec, Pierre Richemond, Elena Buchatskaya, Carl Doersch, Bernardo Avila~Pires, Zhaohan Guo, Mohammad Gheshlaghi~Azar, et~al.
\newblock Bootstrap your own latent-a new approach to self-supervised learning.
\newblock \emph{Advances in Neural Information Processing Systems (NeurIPS)}, 2020.

\bibitem[Hadsell et~al.(2006)Hadsell, Chopra, and LeCun]{hadsell2006dimensionality}
Raia Hadsell, Sumit Chopra, and Yann LeCun.
\newblock Dimensionality reduction by learning an invariant mapping.
\newblock In \emph{Proceedings of IEEE Conference on Computer Vision and Pattern Recognition (CVPR)}, 2006.

\bibitem[He et~al.(2016)He, Zhang, Ren, and Sun]{he2016deep}
Kaiming He, Xiangyu Zhang, Shaoqing Ren, and Jian Sun.
\newblock Deep residual learning for image recognition.
\newblock In \emph{Proceedings of IEEE Conference on Computer Vision and Pattern Recognition (CVPR)}, 2016.

\bibitem[He et~al.(2020)He, Fan, Wu, Xie, and Girshick]{he2020moco}
Kaiming He, Haoqi Fan, Yuxin Wu, Saining Xie, and Ross Girshick.
\newblock Momentum contrast for unsupervised visual representation learning.
\newblock In \emph{Proceedings of the IEEE/CVF Conference on Computer Vision and Pattern Recognition (CVPR)}, pp.\  9729--9738, 2020.

\bibitem[Howard et~al.(2019)Howard, Sandler, Chu, Chen, Chen, Tan, Wang, Zhu, Pang, Vasudevan, Le, and Adam]{howard2017mobilenets}
Andrew Howard, Mark Sandler, Grace Chu, Liang-Chieh Chen, Bo~Chen, Mingxing Tan, Weijun Wang, Yukun Zhu, Ruoming Pang, Vijay Vasudevan, Quoc~V. Le, and Hartwig Adam.
\newblock Searching for mobilenetv3.
\newblock In \emph{Proceedings of IEEE International Conference on Computer Vision (ICCV)}, 2019.

\bibitem[Hudson \& Manning(2019)Hudson and Manning]{hudson2019gqa}
Drew~A Hudson and Christopher~D Manning.
\newblock {GQA}: A new dataset for real-world visual reasoning and compositional question answering.
\newblock In \emph{Proceedings of IEEE Conference on Computer Vision and Pattern Recognition (CVPR)}, 2019.

\bibitem[Jia et~al.(2021)Jia, Yang, Xia, Chen, Parekh, Pham, Le, Sung, Li, and Duerig]{jia2021scaling}
Chao Jia, Yinfei Yang, Ye~Xia, Yi-Ting Chen, Zarana Parekh, Hieu Pham, Quoc Le, Yun-Hsuan Sung, Zhen Li, and Tom Duerig.
\newblock Scaling up visual and vision-language representation learning with noisy text supervision.
\newblock In \emph{Proceedings of the International Conference on Machine Learning (ICML)}, 2021.

\bibitem[Johnson et~al.(2019)Johnson, Douze, and J{\'e}gou]{johnson2019faiss}
Jeff Johnson, Matthijs Douze, and Herv{\'e} J{\'e}gou.
\newblock Billion-scale similarity search with {GPUs}.
\newblock \emph{IEEE Transactions on Big Data}, 2019.

\bibitem[Kazemzadeh et~al.(2014)Kazemzadeh, Ordonez, Matten, and Berg]{kazemzadeh2014referitgame}
Sahar Kazemzadeh, Vicente Ordonez, Mark Matten, and Tamara Berg.
\newblock Referitgame: Referring to objects in photographs of natural scenes.
\newblock In \emph{Proceedings of the Conference on Empirical Methods in Natural Language Processing (EMNLP)}, 2014.

\bibitem[Kornblith et~al.(2019)Kornblith, Shlens, and Le]{kornblith2019better}
Simon Kornblith, Jonathon Shlens, and Quoc~V Le.
\newblock Do better imagenet models transfer better?
\newblock In \emph{Proceedings of the IEEE/CVF conference on computer vision and pattern recognition}, pp.\  2661--2671, 2019.

\bibitem[Lee et~al.(2022)Lee, Kim, Shon, Kim, Kim, Lee, and Kim]{lee2022uniclip}
Janghyeon Lee, Jongsuk Kim, Hyounguk Shon, Bumsoo Kim, Seung~Hwan Kim, Honglak Lee, and Junmo Kim.
\newblock Uniclip: Unified framework for contrastive language-image pre-training.
\newblock In \emph{Advances in Neural Information Processing Systems (NeurIPS)}, 2022.

\bibitem[Li et~al.(2023)Li, Li, Savarese, and Hoi]{li2023blip2}
Junnan Li, Dongxu Li, Silvio Savarese, and Steven Hoi.
\newblock {BLIP-2}: Bootstrapping language-image pre-training with frozen image encoders and large language models.
\newblock In \emph{Proceedings of the International Conference on Machine Learning (ICML)}, 2023.

\bibitem[Li et~al.(2022)Li, Liang, Zhao, Cui, Ouyang, Shao, Yu, and Yan]{li2022declip}
Yangguang Li, Feng Liang, Lichen Zhao, Yufeng Cui, Wanli Ouyang, Jing Shao, Fengwei Yu, and Junjie Yan.
\newblock {Supervision Exists Everywhere: A Data Efficient Contrastive Language-Image Pre-training Paradigm}.
\newblock In \emph{Proceedings of the International Conference on Learning Representations (ICLR)}, 2022.

\bibitem[Loshchilov \& Hutter(2016)Loshchilov and Hutter]{loshchilov2016sgdr}
Ilya Loshchilov and Frank Hutter.
\newblock {SGDR}: Stochastic gradient descent with warm restarts.
\newblock \emph{arXiv preprint arXiv:1608.03983}, 2016.

\bibitem[Mu et~al.(2022)Mu, Kirillov, Wagner, and Xie]{mu2021slip}
Norman Mu, Alexander Kirillov, David Wagner, and Saining Xie.
\newblock Slip: Self-supervision meets language-image pre-training.
\newblock In \emph{Proceedings of European Conference on Computer Vision (ECCV)}, 2022.

\bibitem[Oord et~al.(2018)Oord, Li, and Vinyals]{oord2018representation}
Aaron van~den Oord, Yazhe Li, and Oriol Vinyals.
\newblock Representation learning with contrastive predictive coding.
\newblock \emph{arXiv preprint arXiv:1807.03748}, 2018.

\bibitem[Papineni et~al.(2002)Papineni, Roukos, Ward, and Zhu]{papineni2002bleu}
Kishore Papineni, Salim Roukos, Todd Ward, and Wei-Jing Zhu.
\newblock Bleu: a method for automatic evaluation of machine translation.
\newblock In \emph{Proceedings of the Annual Meeting on Association for Computational Linguistics (ACL)}, 2002.

\bibitem[Park et~al.(2022)Park, Azab, Xiong, Moon, Metze, Kundu, and Ahmed]{park2022normalized}
Yookoon Park, Mahmoud Azab, Bo~Xiong, Seungwhan Moon, Florian Metze, Gourab Kundu, and Kirmani Ahmed.
\newblock Normalized contrastive learning for text-video retrieval.
\newblock \emph{Proceedings of the Conference on Empirical Methods in Natural Language Processing (EMNLP)}, 2022.

\bibitem[Radford et~al.(2021)Radford, Kim, Hallacy, Ramesh, Goh, Agarwal, Sastry, Askell, Mishkin, Clark, et~al.]{radford2021learning}
Alec Radford, Jong~Wook Kim, Chris Hallacy, Aditya Ramesh, Gabriel Goh, Sandhini Agarwal, Girish Sastry, Amanda Askell, Pamela Mishkin, Jack Clark, et~al.
\newblock Learning transferable visual models from natural language supervision.
\newblock In \emph{Proceedings of the International Conference on Machine Learning (ICML)}, 2021.

\bibitem[Reimers \& Gurevych(2019)Reimers and Gurevych]{reimers2019sbert}
Nils Reimers and Iryna Gurevych.
\newblock Sentence-bert: Sentence embeddings using siamese bert-networks.
\newblock In \emph{Proceedings of the 2019 Conference on Empirical Methods in Natural Language Processing}. Association for Computational Linguistics, 11 2019.
\newblock URL \url{https://arxiv.org/abs/1908.10084}.

\bibitem[Selvaraju et~al.(2017)Selvaraju, Cogswell, Das, Vedantam, Parikh, and Batra]{selvaraju2017grad}
Ramprasaath~R Selvaraju, Michael Cogswell, Abhishek Das, Ramakrishna Vedantam, Devi Parikh, and Dhruv Batra.
\newblock Grad-cam: Visual explanations from deep networks via gradient-based localization.
\newblock In \emph{Proceedings of IEEE International Conference on Computer Vision (ICCV)}, 2017.

\bibitem[Sharif~Razavian et~al.(2014)Sharif~Razavian, Azizpour, Sullivan, and Carlsson]{sharif2014cnn}
Ali Sharif~Razavian, Hossein Azizpour, Josephine Sullivan, and Stefan Carlsson.
\newblock Cnn features off-the-shelf: an astounding baseline for recognition.
\newblock In \emph{CVPRW}, 2014.

\bibitem[Suhr et~al.(2019)Suhr, Zhou, Zhang, Zhang, Bai, and Artzi]{suhr2019corpus}
Alane Suhr, Stephanie Zhou, Ally Zhang, Iris Zhang, Huajun Bai, and Yoav Artzi.
\newblock A corpus for reasoning about natural language grounded in photographs.
\newblock In \emph{Proceedings of the Annual Meeting on Association for Computational Linguistics (ACL)}, 2019.

\bibitem[Vedantam et~al.(2015)Vedantam, Lawrence~Zitnick, and Parikh]{vedantam2015cider}
Ramakrishna Vedantam, C~Lawrence~Zitnick, and Devi Parikh.
\newblock {CIDEr: Consensus-based image description evaluation}.
\newblock In \emph{Proceedings of IEEE Conference on Computer Vision and Pattern Recognition (CVPR)}, 2015.

\bibitem[Vincent et~al.(2008)Vincent, Larochelle, Bengio, and Manzagol]{vincent2008extracting}
Pascal Vincent, Hugo Larochelle, Yoshua Bengio, and Pierre-Antoine Manzagol.
\newblock Extracting and composing robust features with denoising autoencoders.
\newblock In \emph{Proceedings of the International Conference on Machine Learning (ICML)}, 2008.

\bibitem[Wang et~al.(2019)Wang, Chao, Weinberger, and van~der Maaten]{wang2019simpleshot}
Yan Wang, Wei-Lun Chao, Kilian~Q. Weinberger, and Laurens van~der Maaten.
\newblock Simpleshot: Revisiting nearest-neighbor classification for few-shot learning.
\newblock \emph{arXiv preprint arXiv:1911.04623}, 2019.

\bibitem[Wightman et~al.(2021)Wightman, Touvron, and Jegou]{wightman2021resnet}
Ross Wightman, Hugo Touvron, and Herve Jegou.
\newblock Resnet strikes back: An improved training procedure in timm.
\newblock In \emph{NeurIPS Workshop on ImageNet: Past, Present, and Future}, 2021.

\bibitem[Wu et~al.(2018)Wu, Xiong, Yu, and Lin]{wu2018unsupervised}
Zhirong Wu, Yuanjun Xiong, Stella~X Yu, and Dahua Lin.
\newblock Unsupervised feature learning via non-parametric instance discrimination.
\newblock In \emph{Proceedings of the IEEE conference on computer vision and pattern recognition}, pp.\  3733--3742, 2018.

\bibitem[Xu et~al.(2022)Xu, De~Mello, Liu, Byeon, Breuel, Kautz, and Wang]{xu2022groupvit}
Jiarui Xu, Shalini De~Mello, Sifei Liu, Wonmin Byeon, Thomas Breuel, Jan Kautz, and Xiaolong Wang.
\newblock Groupvit: Semantic segmentation emerges from text supervision.
\newblock In \emph{Proceedings of IEEE Conference on Computer Vision and Pattern Recognition (CVPR)}, 2022.

\bibitem[Yao et~al.(2022)Yao, Huang, Hou, Lu, Niu, Xu, Liang, Li, Jiang, and Xu]{yao2022filip}
Lewei Yao, Runhui Huang, Lu~Hou, Guansong Lu, Minzhe Niu, Hang Xu, Xiaodan Liang, Zhenguo Li, Xin Jiang, and Chunjing Xu.
\newblock {FILIP}: Fine-grained interactive language-image pre-training.
\newblock In \emph{Proceedings of the International Conference on Learning Representations (ICLR)}, 2022.

\bibitem[Young et~al.(2014)Young, Lai, Hodosh, and Hockenmaier]{young2014image}
Peter Young, Alice Lai, Micah Hodosh, and Julia Hockenmaier.
\newblock From image descriptions to visual denotations: New similarity metrics for semantic inference over event descriptions.
\newblock \emph{TACL}, 2014.

\bibitem[Zellers et~al.(2019)Zellers, Bisk, Farhadi, and Choi]{zellers2019recognition}
Rowan Zellers, Yonatan Bisk, Ali Farhadi, and Yejin Choi.
\newblock From recognition to cognition: Visual commonsense reasoning.
\newblock In \emph{Proceedings of IEEE Conference on Computer Vision and Pattern Recognition (CVPR)}, 2019.

\bibitem[Zhang et~al.(2016)Zhang, Isola, and Efros]{zhang2016colorful}
Richard Zhang, Phillip Isola, and Alexei~A Efros.
\newblock Colorful image colorization.
\newblock In \emph{Proceedings of European Conference on Computer Vision (ECCV)}, 2016.

\bibitem[Zheng et~al.(2021)Zheng, Wang, You, Qian, Zhang, Wang, and Xu]{zheng2021weakly}
Mingkai Zheng, Fei Wang, Shan You, Chen Qian, Changshui Zhang, Xiaogang Wang, and Chang Xu.
\newblock Weakly supervised contrastive learning.
\newblock In \emph{Proceedings of IEEE International Conference on Computer Vision (ICCV)}, 2021.

\bibitem[Zhu et~al.(2016)Zhu, Groth, Bernstein, and Fei-Fei]{zhu2016visual7w}
Yuke Zhu, Oliver Groth, Michael Bernstein, and Li~Fei-Fei.
\newblock Visual7w: Grounded question answering in images.
\newblock In \emph{Proceedings of IEEE Conference on Computer Vision and Pattern Recognition (CVPR)}, 2016.

\end{thebibliography}
\bibliographystyle{tmlr}

% \appendix
% \section{Appendix}
% You may include other additional sections here.

\end{document}